\documentclass[conference]{IEEEtran}
\IEEEoverridecommandlockouts

\usepackage[english]{babel}
\usepackage{soul}
\usepackage[dvipsnames]{xcolor}
\usepackage[colorlinks=true, linkcolor=MidnightBlue, citecolor=MidnightBlue, urlcolor=MidnightBlue]{hyperref}
\usepackage[ruled,vlined,linesnumbered]{algorithm2e}
\SetKwInput{KwInput}{Input}                
\SetKwInput{KwOutput}{Output}              
\SetKwInput{KwInitalise}{Initialise}
\usepackage{tikz}			
\usepackage{pgfplots}		
\usetikzlibrary{calc,arrows.meta,positioning}
\usepackage{booktabs}       
\usepackage{makecell} 
\usepackage{adjustbox}  
\usepackage{float}  
\restylefloat{table}

\usepackage{fancyhdr}
\usepackage{cite}
\usepackage{amsmath,amssymb,amsfonts}
\usepackage{algorithmic}
\usepackage{graphicx}
\usepackage{textcomp}
\usepackage{xcolor}
\def\BibTeX{{\rm B\kern-.05em{\sc i\kern-.025em b}\kern-.08em
    T\kern-.1667em\lower.7ex\hbox{E}\kern-.125emX}}

\DeclareMathOperator*{\argmax}{argmax} 

\begin{document}

\title{Bootstrapped model learning and error correction for planning with uncertainty in model-based RL
\thanks{A diagram of the architecture, list of hyperparameters and source code are located in \url{https://github.com/aovalle/bootstrapped-transition-functions}}
}

\author{\IEEEauthorblockN{Alvaro Ovalle}
\IEEEauthorblockA{\textit{School of Electronic Engineering and Computer Science} \\
\textit{Queen Mary University of London}\\
London, UK \\
a.ovalle@qmul.ac.uk}
\and
\IEEEauthorblockN{Simon M. Lucas}
\IEEEauthorblockA{\textit{School of Electronic Engineering and Computer Science} \\
\textit{Queen Mary University of London}\\
London, UK \\
simon.lucas@qmul.ac.uk}
}

\maketitle

\begin{abstract}

Having access to a forward model enables the use of planning algorithms such as Monte Carlo Tree Search and Rolling Horizon Evolution. Where a model is unavailable, a natural aim is to learn a model that reflects accurately the dynamics of the environment. In many situations it might not be possible and minimal glitches in the model may lead to poor performance and failure. This paper explores the problem of model misspecification through uncertainty-aware reinforcement learning agents. We propose a bootstrapped multi-headed neural network that learns the distribution of future states and rewards. We experiment with a number of schemes to extract the most likely predictions. Moreover, we also introduce a global error correction filter that applies high-level constraints guided by the context provided through the predictive distribution. We illustrate our approach on Minipacman. The evaluation demonstrates that when dealing with imperfect models, our methods exhibit increased performance and stability, both in terms of model accuracy and in its use within a planning algorithm.
\end{abstract}


\section{Introduction}

In recent years model-based reinforcement learning (RL) has experienced a revival boosted by advances in deep learning. Compared to a reactive strategy, possessing an internal model (i.e. forward model) and being able to encode knowledge about the dynamics of the environment offers several advantages. Having a model appears to be a key component enabling the capacity to forecast events, anticipate consequences, evaluate possibilities or visualise different perspectives. 

Beyond these positive aspects, model-based RL could also potentially lessen some of the technical issues that afflict model-free approaches, such as sample inefficiency and lack of generalisation. Ideally if an agent acquires a model, it gains the ability to simulate experiences to plan or to learn a policy without the necessity of interacting directly with the environment. This implies a considerable advantage in robotic control tasks or those requiring safe exploration. Furthermore, by encoding environmental dynamics the knowledge can be reused and leveraged for different tasks.

However acquiring a forward model that is accurate enough to support planning comes with its own set of challenges. The agent does not only have to learn a policy as in model-free RL, but also a model on which to learn it, introducing an additional source of error. Crucially since the model approximates the dynamics, even small errors can be catastrophic, making the model essentially useless as they accumulate over the rollout. Nonetheless, very promising research has focused on how to make models more reliable through various techniques. For example by learning and executing them completely in latent space \cite{buesing2018,hafner2019}, by adding noise to increase their robustness \cite{ha2018} or through a combination of multiple techniques \cite{asadi2019}. Alternative approaches sideline learning state transitions altogether, and focus instead on modelling other aspects such as value functions and future policy \cite{schrittwieser2020}. Despite these advances it is almost unavoidable, that for any complex task, a model will be inherently incomplete. There will be an epistemic component related to a gap in knowledge that could decrease by further learning or new data. However there will also be an aleatoric aspect associated to the stochasticity of the process that generates the data itself \cite{kendall2017}.

In this paper we are concerned with the importance it has for an agent to have access to its uncertainty. Different from \cite{buesing2018}, we are not specifically interested in finding better state representation mechanisms intended to learn a more accurate model. Instead it is assumed the agent must deal with the faulty or premature models it currently has, and yet, it must plan and act under uncertainty. The method we propose employs \textit{bootstrapping} techniques to approximate the predictive distribution of state transitions by learning an ensemble of models through a single multi-headed architecture. We devise and test three different schemes to integrate the data from the distribution. In this sense our objective is closer to that pursued in \cite{racaniere2017}, where the goal is to assist the decision making process by aggregating information from a group of inaccurate models. In addition we also try to demonstrate how having access to a predictive distribution enables the construction of error-correcting routines, increasing the reliability of the predictions. Concretely, we consider an environment with multiple interacting elements, whose presence or absence follows a consistent set of rules serving as constraints. If the initial prediction of an agent does not satisfy those constraints, alternative predictions contained in the distribution can inform how to adjust the original prediction. We test our approach on a video game control task where it exhibits increased performance compared to a non-probabilistic agent.

\section{Background and Previous Work}\label{section:background}

\subsection{Model-Based Reinforcement Learning}

In RL we denote state $s_t \in \mathcal{S}$, action $a_t \in \mathcal{A}$ and a reward function $r(s_t,a_t)$. The environment is governed by a transition function or model $f(s_t,a_t)=s_{t+1}$ providing the next state. Thus in the model-based case a mapping to $f$, which we refer to as $f_\theta$, is either provided or learned from observations. For stochastic environments $f(s_{t+1}|s_t, a_t)=Pr(s_{t+1}|s_t,a_t)$. It is possible to express a simulated rollout of action-state sequences $\tau=(s_1,a_1, \dots, s_T,a_T)$ with respect to a policy $\pi$ as a distribution:

\begin{equation}
    p_{\theta,\phi}(\tau) = p_\theta(s_1) \prod_t \pi_{\phi}(a_t|s_t)p_\theta(s_{t+1}|s_t,a_t)
\end{equation} 

The goal of a model-based RL agent is to find a policy $\pi$ that maximises the rewards along this simulated trajectory subjected to $\max_{\theta,\phi} \mathbb{E}_{\tau \sim p_{\theta,\phi}} \big[ \sum_t r(s_t,a_t) \big]$.

\subsection{The Bootstrap Method}\label{section:bootstrap}

The bootstrap method is an statistical technique to assess the accuracy of an \textit{statistic} or \textit{estimator} of an unknown population parameter \cite{efron1979}. In some cases the variability of the estimator of interest is established by repeatedly sampling from the population itself. However, sometimes this may not be feasible and the only measurements at our disposal come from a single observed pool of sample data $D$. The underlying principle behind the bootstrap method relies on the assumption that the sample data is representative of the true population. Thus we can simulate the process of randomly sampling from the population by treating the sample data as if it were the population. If the number of samples is large enough, it can be possible to obtain a distribution that approximates the sampling distribution of the estimator. The process consists of the following steps:

\begin{itemize}
	\item From a sample data $D$ of size $N$ resample with replacement to obtain a bootstrap sample $\tilde{D}$ of the same size. This process is repeated to get $K$ bootstrap samples.
	\item Compute an \textit{statistic} or \textit{estimator} $\chi^*_k$ for each of the bootstrap samples.
	\item Build a bootstrap distribution with the $K$ estimators.
\end{itemize}

From the bootstrap distribution then it is possible to perform inference. For example, by calculating confidence intervals or obtaining the standard error to assess the variability across estimates. 

\begin{figure}
    \centering
    \scalebox{0.75}{



\begin{tikzpicture}

\node[draw, circle, very thick, label={[label distance=2cm, align=center]above:Sample\\Data}, pin={[fill=gray!20, rounded corners=5pt, align=center]below:Estimator\\$\chi_0$}, matrix,inner color=gray, outer color=yellow!20!gray](sample){
	\foreach \p in {1,...,50}
	{ \fill[yellow] (0.5*rand,0.5*rand) circle (0.05);
	};\\
};

\node[draw, circle, right= of sample, pin={[name=lbs2, pin distance=10mm, fill=gray!20, minimum width=0pt, align=left, rounded corners=5pt]right:{$\chi^*_2$}}, matrix,inner color=gray, outer color=yellow!20!gray](bs2){
	\foreach \p in {1,...,50}
	{ \fill[yellow] (0.5*rand,0.5*rand) circle (0.05);
	};\\
};

\node[draw, circle, above= of bs2, pin={[name=lbs1, pin distance=10mm, fill=gray!20, minimum width=0pt, align=left, rounded corners=5pt]right:{$\chi^*_1$}}, matrix,inner color=gray, outer color=yellow!20!gray](bs1){
	\foreach \p in {1,...,50}
	{ \fill[yellow] (0.5*rand,0.5*rand) circle (0.05);
	};\\
};

\node[draw, circle, below = of bs2, pin={[name=lbsk, pin distance=10mm, fill=gray!20, minimum width=0pt, align=left, rounded corners=5pt]right:{$\chi^*_k$}}, matrix,inner color=gray, outer color=yellow!20!gray](bsk){
	\foreach \p in {1,...,50}
	{ \fill[yellow] (0.5*rand,0.5*rand) circle (0.05);
	};\\
};

\draw[->, very thick, out=90,in=-90] (sample.10) to node[auto] {} (bs1.south);
\draw[->, very thick] (sample) to node[auto] {} (bs2);
\draw[->, very thick, out=-90,in=90] (sample.-10) to node[auto] {} (bsk);

\node at ($(lbs2)!.5!(lbsk)$) {\vdots};

\node[rectangle, fill=blue!20!gray, draw, scale=0.3, minimum size=20em,right = of lbs2]  (gauss) {
	\begin{tikzpicture}
	\begin{axis}[axis lines=none, ticks=none,xmax=3, xmin=-3,ymax=1.1]
	\addplot[ultra thick,white, no markers,samples=200] {exp(-x^2)};
	\end{axis}
	\end{tikzpicture}
};

\draw[->, out=0,in=90] (lbs1.east) to node[auto] {} (gauss.north);
\draw[->] (lbs2.east) to node[auto] {} (gauss);
\draw[->, out=0,in=-90] (lbsk.east) to node[auto] {} (gauss.south);

\node[above=0.3cm of bs1, align=center](bootlabel) {Bootstrap\\Samples};
\node[right=10pt of bootlabel, align=center](estlabel) {Bootstrap\\Estimators};
\node[above=1.8cm of gauss, align=center](estlabel) {Bootstrap\\Distribution};

\end{tikzpicture}

}
    \caption{The bootstrap method. The process consists in resampling with replacement from an original group of observations. For each new sample pool an estimator is computed and then used to build the bootstap distribution.} \label{fig:bootstrapping}
\end{figure}
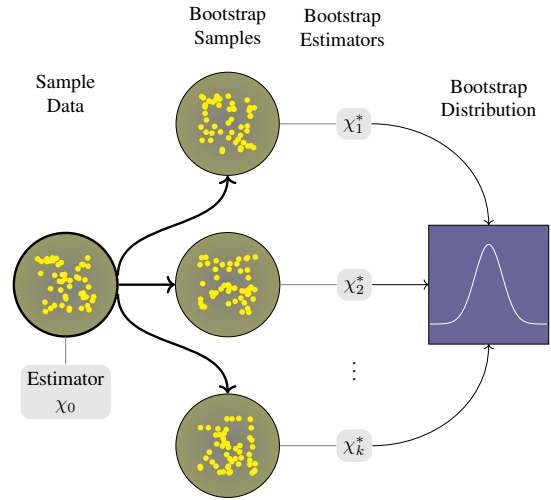

\subsection{Bootstrapping in Reinforcement Learning and Randomised Prior Functions}\label{section:bootdqn}

The idea behind the bootstrap method has found application in RL to estimate uncertainty and to facilitate more principled and efficient exploration strategies. \cite{osband2016a} proposed the \textit{bootstrapped DQN} to approximate the posterior of a Q-function. Compared to the original DQN \cite{mnih2013} they introduced a multi-headed architecture where each head is trained on a different group of experiences from a replay buffer as determined by a masking distribution. Therefore each head models its own approximation of the state-action values. Initially during training due to the discrepancy in the approximations, the agent chooses different actions depending on the head doing the decision making, which drives exploration in a natural manner. As training progresses the heads tend to converge to an estimate. After training, the agent can assess its confidence about the state-action value of an observation by sending it through each head and obtaining $K$ different predictions (i.e. the bootstrap distribution)

A shortcoming of the bootstrap method is that by itself it does not include a mechanism to reflect uncertainty beyond what is coming from the observed data. The authors pointed out that this issue is mitigated by relying on the diversity that is induced by the random initialisation of the network which acts as a form of prior. However this is further addressed in \cite{osband2018} with the inclusion of randomised prior functions (RPF). The approach consists in explicitly specifying a prior via an additional fixed and randomly initialised network. This can be described as:

\begin{equation}
    Q_{\theta_k}(x) = f_{\theta_k}(x) + \beta p_k
\end{equation}

Where $Q_{\theta_k}$ corresponds to head $k$ and it is modelled by a function approximator $f_{\theta_k}$ with tunable parameters $\theta_k$ and a non-trainable prior $p_k$ scaled by $\beta$.

The bootstrapped DQN has also been extended to model-based RL \cite{huang2019}, helping in the identification of state-action values with a larger level of uncertainty to guide search in look-ahead trees \cite{farquhar2018}.

\section{Method}

\subsection{Bootstrapped Transition Functions} \label{section:btf}

In this work we also take the idea behind bootstrapped DQN, but instead of applying it to Q-functions we turn our interest towards environment transitions. Thus we take a probabilistic interpretation to the problem of estimating the next set of future observations and rewards. 

Formally we can represent a neural network as a conditional model $p_\theta(y|x)$ parameterised by $\theta$, where $x$ is the input and $y$ is the output. In the RL context, consider an agent that learns $p_\theta(s_{t+1}, r_t |s_t, a_t)$ as it observes data $D=\{s_i,a_i,r_i,s_{i+1}\}_i^N$. That is, it learns the set of assumptions of the model, encoded in the parameters $\theta$, that explains $D$ as good as possible. The posterior distribution $p(\theta|D)$ is usually intractable and often the parameters $\theta$ are learned as point estimates by calculating the maximum likelihood or the maximum a posteriori when introducing some form of regularisation. 

If we require to work with a distribution, a popular approach consists in finding the network parameters $\phi$ of a candidate distribution $q$ that is as similar as possible to the true posterior distribution (i.e. $q_\phi(\theta|D)\approx p(\theta|D)$) using variational inference to minimise $D_{KL}[q_\phi(\theta|D)||p(\theta|D) ]$ \cite{hinton1993,graves2011,blundell2015}. In our method instead of learning $q_{\phi}$, we assume that the parameters on each head $k$, plus those in the part of the network that is shared amongst the heads, are a sample $\theta_k \sim p(\theta|D)$. Together all these samples form an ensemble distribution $q_e$ that approximates the true posterior $p$. Then we can use $q_e$ to obtain an estimate of the predictive distribution of next state $s_{t+1}$ and reward $r_t$ given a new state-action input:

\begin{equation}
p(s_{t+1}, r_t|s_{t}, a_t) \approx \mathbb{E}_{\theta \sim q_e(\theta|D)}[p_\theta(s_{t+1}, r_t|s_{t}, a_t)]
\end{equation}

The predictive distribution corresponds to the bootstrap distribution reviewed in sections \ref{section:bootstrap} and \ref{section:bootdqn}. Every head outputs a prediction of the next state and reward and can be interpreted as estimators $\chi^*_{s'_i,r_i}$. From this distribution we can quantify the uncertainty and compute other statistical metrics of interest. 

Any behavioural policy can be followed to learn the multi-headed environment models. To increase the stability and generalisation, the models are not learned online but from a minibatch sampled from a replay buffer and trained to minimise the cross-entropy loss between predictions and actual observations or rewards. In addition to vanilla bootstrapping of transition functions, we were also interested in comparing the effects of adding RPFs. As per equation 1 when a prior was included the predicted next state an reward on head $k$ corresponded respectively to:

\begin{align}
    \begin{split}
T_{\theta_k} (s_t,a_t) = f^t_{\theta_k} (s_t,a_t) + \beta p_k
\\
r_{\theta_k} (s_t,a_t) = f^r_{\theta_k} (s_t,a_t) + \beta p_k
    \end{split}
\end{align}

\begin{algorithm}
    \SetAlgoLined
    \KwInput{Policy $\pi$, number of heads $K$}
    \KwInitalise{Replay buffer $D$, parameters $\theta$}
    \For{$t \dots T$}{
        Sample $a_t \sim \pi(a_t|s_t)$\;
        Observe $s_{t+1}, r_t \sim p(s_{t+1},r_t|s_t, a_t)$\;
        \If{done}{
            $env.reset()$\;
        }
        Generate masks $m \sim Ber(K, 0.5)$\;
        $D \leftarrow D\cup \{s,a,r,s',m\}$\;
        Sample $s_m, a_m, r_m, s'_m, m_m \sim D$\;
        $\hat{y} = \{\hat{s}_{t+1}, \hat{r}_t\} = f_\theta(s_m, a_m)$\;
        Update model by minimising $\mathcal{L}_\theta(\hat{y},y)$, backpropagating according to $m$ via SGD\;
    }
    
    \caption{Bootstrapped Transition Functions}
\end{algorithm}

\subsection{Rollout strategy}

Regardless of the level of accuracy contained in a learned model of the environment. An agent harnesses it to simulate spatio-temporal prospects and contingencies. We use Rolling Horizon Evolution (RHE)\cite{perez2013} for integrating the knowledge encoded in the model with the capacity to manipulate it in order to imagine potential outcomes. 

RHE is a real-time control and planning algorithm that uses a forward model to search the space of trajectories in order to maximise certain utility. We can define a policy $\pi$ as an action sequence $a_{0:\mathcal{T}}$ of length $\mathcal{T}$, starting from a current state $s_0$ to $s_\mathcal{T}$ (or an earlier terminal state). Thus RHE attempts to find a policy that satisfies, within certain budget, the following:

\begin{equation}
    \pi(s) = \argmax_{a_{0:\mathcal{T}}} \mathbb{E}\Bigg[ r(s_0, a_0) + \sum^{\mathcal{T}}_{t=1}r(\hat{s}_t, a_t) \Bigg]
\end{equation}

Where $r(s,a)$ is the reward function. It has to be noted that only $s_0$ is an actual observed state while $\hat{s}_t$ are states simulated via a forward model. 

A major advantage of RHE is its simplicity while retaining competitive performance against other well established planning algorithms such as MCTS \cite{perez2013, liu2016, santos2018}. In its most basic formulation, RHE works by mutating an initial random action sequence to get a population of size $\mathcal{P}$. The fitness of the action sequences is determined by evaluating them using the forward model. The action sequence with the highest total reward in a rollout is selected, and the first action of the sequence is carried out in the real environment. The process is described in algorithm \ref{algo:rh}.

Several enhancements intended to improve the performance of RHE have been proposed \cite{gaina2017,gaina2017a,tong2019}. For the moment we focus on the vanilla version of RHE that we have just briefly summarised. The only enhancement we have added to this standard version is the \textit{shift buffer} technique which aims to facilitate information transfer from the past \cite{gaina2017}. This consists in seeding the new population with the previous fittest action sequence, by shifting it one time step to the left and appending a new random action at the end of the sequence to preserve its length.

RHE has traditionally been applied to perfect simulators but the basic mechanics of the algorithm remain unchanged when using imperfect forward models. However one must establish how to obtain the imagined states $\hat{s}$ on which RHE will operate. Taking into account that the network has $k$ heads and that each of them generates its own predictions, it is necessary to determine how to tackle the discrepancy arising from the uncertainty of the forward model to consolidate the outputs into a single unified prediction. We considered three different criteria for next state and reward prediction. 
\bigskip

\begin{itemize}
\item \textbf{Average:} computes the mean of the output values in the last layer of each head and retrieves the class with the largest value:

\begin{equation}
    \label{eqn:avg}
    \hat{s}_{t+1}, \hat{r}_t = \argmax_c \Bigg[\dfrac{1}{K} \sum_k \sigma\Big(\mathbf{u}^{(L)} \Big)_k \Bigg]_c    
\end{equation}

Where $\mathbf{u}^{(L)}$ is the vector of logits in the last layer $L$ of head $k$ and $c \in \{0, \dots, C\}$.

\vspace{5mm}

\item \textbf{Majority Voting:} takes the most common prediction by obtaining the mode over the output of each head:

\begin{equation}
    \label{eqn:vote}
\hat{s}_{t+1}, \hat{r}_t = mode\big(f_{\theta_1}, \dots, f_{\theta_K}\big)
\end{equation}

\vspace{4mm}

\item \textbf{Sampling:} selects a prediction from the predictive distribution formed by the output of each head:

\begin{equation}
    \label{eqn:sample}
\hat{s}_{t+1}, \hat{r}_t \sim p\big(s_{t+1},r_t|s_t,a_t;[\theta_k]_{k=1}^K\big)
\end{equation}

\end{itemize}

\begin{algorithm}
    \label{algo:rh}
    \SetAlgoLined
    \KwInput{Forward model $p_\theta$, number of actions $A$, sequence length $\mathcal{T}$, population size $\mathcal{P}$, mutation rate $\mu$}
    \While{true}{
    \eIf{not shift buffer}{
        $\pi \leftarrow a_t, \dots, a_{\mathcal{T}} \sim Cat(A)$\;
    }{
        Shift $\pi$ to the left and add $a_{\mathcal{T}} \sim Cat(A)$ at the end of the sequence\;
    }
    \For{$i \dots \mathcal{P}$}{
        \If{$i > 0$}{$\pi_i \leftarrow$ Mutate $\pi$ with rate $\mu$}
        \For{$h \dots \mathcal{T}$}{
            Predict $\hat{s}_{h+1}$ and $\hat{r}_h$ using eq. \ref{eqn:avg}, \ref{eqn:vote} or \ref{eqn:sample} via $p_{\theta}$\;
            \If{error-correction}{
                $\hat{s}_{h+1} \leftarrow correct(\hat{s}_{h+1})$\;
            } 
            Save or update current return $R(\pi_i^{(0:h)}) = r(s_t, a_0) + \sum^{h}_{k=1}r(\hat{s}_k, a_k)$\;
        }
    }
    Select fittest action sequence $\pi \leftarrow max_{\pi} R(\pi)$\;
    Perform first action in the sequence $a \leftarrow \pi^{(0)}$\;
    Observe $s_{t+1}, r_t \sim p(s_{t+1},r_t|s_t, a)$\;
    }
    \caption{Rolling Horizon Evolution with inaccurate models and error-correction}
\end{algorithm}

\subsection{Error-Correction}\label{section:ec}

It is widely acknowledged that learning a forward model is challenging in complex and noisy tasks. Compound error may render a model impractical for planning. However it is less understood the different ways the forward models can be wrong in their predictions. There is currently no comprehensive classification on the type of errors we may encounter.

A particular type of error is the lack of alignment between predictions done locally and how they scale up at a higher level. Even if local predictions are sensible, it does not necessarily imply they will be consistent once all of them are integrated as a whole. For instance in visual domains where a frame can be modelled by several distributions operating at lower levels such as pixels or objects. Consider the example of a video game frame where a neural network outputs predictions for each section of the frame. The network learns about the existence of various objects in the environment, for example as clouds, trees or animals. If the network predicts a cloud at a given frame section, this prediction by itself and in isolation, might not seem surprising. However if we look at the frame and observe a cloud displayed below a tree, then the predicted frame has a low likelihood of actually occurring in the environment. Simulating further into the future by building upon inaccurate predictions may lead to quick degeneration. Thus in some cases it could be useful to think in terms of the higher level constraints that need to be satisfied by a prediction. 

In this work we focus on two kinds of constraints intended to regulate the amount of objects of certain class that should be present in the predicted frame. Furthermore here we describe how the ensemble can leverage the uncertainty in the estimates via the bootstrap distribution in order to make predictions that are more robust, reliable and that have a higher chance of conforming to the expectations of what is realistically possible in an environment. 

We start from the assumption that some of these constraints are discrete and known to the agent at the beginning of a simulated rollout. Any correction made to the frames is based on what is immediately accessible to the agent: (1) its current observation and/or (2) its predictions. For the latter, having a bootstrap distribution becomes crucial for error-correcting. By having not only a single point estimate but several candidate predictions, we can establish mechanisms that search, compare and integrate observations gathered from the heads into new revised predictions that are more likely to satisfy the criteria demanded by the constraints. Note that we will refer to an \textit{element} as any generic section in the frame such as a pixel, a cell or an object. The two error-correction mechanisms are the following: 

\begin{algorithm}
    \label{algo:ec-missing}
    \SetAlgoLined
    \KwInput{Unified frame prediction $f_u$, Multi-headed predictions $f_{\theta_k}$, element $e$, constraint $c$, last observed or imagined position $lp$}
    \If{$e$ in $f_u < c$}{
        Find positions in each $f_{\theta_k}$ where $e$ is and store in $\textbf{p}$\;
        \eIf{\textbf{p} is not empty}{
            \uIf{sampling}{
                $p' \sim \textbf{p}$\;
            }\ElseIf{average or voting}{
                $p' \leftarrow mode(p_0, \dots, p_N)$\;
            }
        }{
            $p' \leftarrow lp$\;
        }
        Insert $e$ in new position $p'$ in unified frame $f_u$\;
        $lp \leftarrow p'$\;
    }
    \caption{Error-correction for missing elements}
\end{algorithm}

\bigskip

\begin{itemize}
\item \textbf{Missing element:} this type of verification applies to situations where the unified frame prediction does not include the presence of a particular element in any of the cells even though it would be expected. To correct the frame, every head is inspected for the element. If a head has one or multiple predictions containing the element, the positions where the element is located are stored in a shared vector along with the locations coming from other heads. After all heads have been examined, the frame is corrected by inserting the missing element in a position selected from the shared location vector. The selection is done either by taking the mode of the location vector or by sampling from it, depending on the criteria used by the network to predict next state and reward. If the location vector is empty because none of the heads predict the existence of the element then the position is taken from the previous frame. Only for the first step in a simulated rollout, the previous frame really corresponds to the last observation gathered from the environment whereas for the rest of the steps it is taken from the previously imagined frame. The pseudo-code of this mechanism is outlined in algorithm \ref{algo:ec-missing}.

\bigskip

\item \textbf{Additional element:} the other situation considered here is when the prediction includes more elements of a certain type than those anticipated. First we find the locations where the element is present in the unified predicted frame and store them in a vector $\textbf{a}$. Then we locate the positions of the element in every head of the ensemble and store them in a shared vector $\textbf{p}$. Similar to the unified frame prediction, a head can also be predicting the existence of the element more times than those specified by the constraints. The two vectors are then compared to identify and separate the locations in the shared vector $\textbf{p}$ that occur in $\textbf{a}$ as well. From this resulting vector we select the position of the elements that will be preserved in the unified prediction. The selection is done by either taking the mode or by sampling from the vector. 

After it has been decided what elements are preserved, then it has to be determined what should substitute those that have not been chosen. This substitution can also be assisted by the predictions contained in the heads. First by gathering the location of those elements in the unified frame and then by extracting the predictions done by each head in those locations, but considering only the predictions that are different from the element that should be excluded. Once they have been gathered the selection is again made computing the mode or by sampling one of the elements. Similar to the \textit{missing element} routine when a decision cannot be made due to an empty vector, the rectification will default to the element occurring in the last frame, whether real or imagined.

\end{itemize}

\begin{algorithm}
    \label{algo:ec-additional}
    \SetAlgoLined
    \KwInput{Unified frame prediction $f_u$, Multi-headed predictions $f_{\theta_k}$, element $e$, constraint $c$, last imagined or observed element $le$}
    \KwInitalise{Vector \textbf{v} that will hold the elements that replace the copies}
    \If{$e$ in $f_u > c$}{
        Find positions in $f_u$ where $e$ is and store in $\textbf{a}$\;
        Find positions in each $f_{\theta_k}$ where $e$ is and store in $\textbf{p}$\;
        Check which values in $\textbf{a}$ are in $\textbf{p}$ and store in $\textbf{g}$\;
        \If{\textbf{g} is not empty}{
            \uIf{sampling}{
                $g' \sim \textbf{g}$\;
            }\ElseIf{average or voting}{
                $g' \leftarrow mode(g_0, \dots, g_N)$\;
            }
            Remove $g'$ from $\textbf{g}$\;
            \For{$g_i \in \textbf{g}$}{
                For each $f_{\theta_k}$ extract the element located at $g_i$ and store in $\textbf{v}$\;
                \eIf{\textbf{v} is not empty}{
                    \uIf{sampling}{
                        $v' \sim \textbf{v}$\;
                    }\ElseIf{average or voting}{
                        $v' \leftarrow mode(v_0, \dots, v_N)$\;
                    }
                }{
                    $v' \leftarrow le$
                }
                Insert $v'$ in position $g_i$ in unified frame $f_u$\;
                $le \leftarrow v'$\;
            }
        }
    }
    \caption{Error-correction for additional elements}
\end{algorithm}

\section{Experimental Setup}

\begin{figure}[h]
    \includegraphics[scale=0.4]{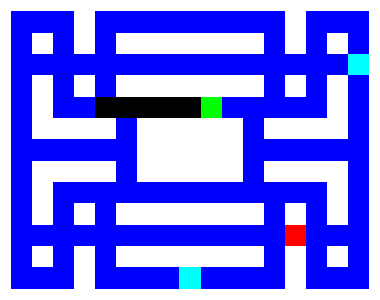}
    \centering
    \caption{Minipacman. The green cell corresponds to the agent, red to the ghost, turquoise to the power pills, blue to the food still left in the maze and black to a cell where the food has been eaten.} \label{fig:minipacman}
\end{figure}

\subsection{Environment}

We demonstrate the effects of the bootstrapped transition functions with and without error correction mechanisms in Minipacman \footnote{Our version has been modified and adapted to the task. It was originally taken from \url{https://github.com/higgsfield/Imagination-Augmented-Agents}} \cite{racaniere2017} (fig. \ref{fig:minipacman}). The environment is a reduced version of Pacman and provides a simple minimal discrete control benchmark. The game consists in an agent that navigates through a maze eating food in the corridors. The agent is chased by a ghost moving semi-randomly. This element of stochasticity arising from the behaviour of the ghost is an additional challenge for learning a forward model, as it deals not only with the epistemic factors of the environment but also the aleatoric. In addition, the game contains power pills that when eaten by the agent allows it for a fixed number of steps the possibility to eat the ghost. The agent has a repertoire of five actions: up, down, left, right and no operation. It also receives a reward depending on specific game events. When the agent eats food it gets a reward of 1, if it does not move or goes to a section of the corridor with no food it does not receive a reward. It gets a reward of 3 for the power pills and 6 if it manages to eat the ghost whilst under the effects of the power pills. If the ghost kills the agent the reward is -1 and the game instance terminates.

Besides the reward, the agent also receives an observation that reflects the current state of the game. The format of the observation is an array of $15 \times 19\times 3$, where the last number is the amount of channels. Thus the observation contains a total of $15 \times 19$ cells. Each of them is associated with a different object either an agent, food, power pills, a ghost, eaten and inaccessible cells. The fact that we can treat a cell as an object allows us to evaluate our error-correcting strategy as proof of concept, since unsupervised object detection for control tasks is still an nascent area of research \cite{kulkarni2019, watters2019, anand2020}.

\subsection{Error-Correction in Minipacman}

The algorithms described in section \ref{section:ec} offer a general strategy. A starting point for error-correcting a simulated rollout. To test their performance we focused on the most fundamental elements of the game: the agent and the ghost. Predicted frames are checked against the constraints established for the game. The constraints are simply the number of agents or ghosts counted from the last real observation (or the last imagined observation). If there are less or more than the number of elements expected, the routines in algorithm \ref{algo:ec-missing} and algorithm \ref{algo:ec-additional} are invoked respectively. There is however a small element-specific adjustment done for algorithm \ref{algo:ec-missing}. If the predicted frame initially contains more than a single pacman, then from the pool of copies to be discarded, we verify if one of them corresponds to the position where it was observed in the last frame. If the conditon is met then the cell is turned black, symbolizing that the food there has been eaten. Otherwise they follow the original outline and take their values from the last observed or imagined frame.

\subsection{Architecture}

For learning the forward model of the environment we implemented a convolutional neural network (CNN) with an additional fully connected last layer. The input consists of a concatenation of the last frame and a broadcasted one-hot representation of the previous action. The CNN attempts to capture long range temporal dependencies through \textit{pool-and-inject} layers (we refer the reader to \cite{racaniere2017} for more details). The output of the network is the categorical representation of every cell in the next predicted frame and the predicted reward. The frame and the reward are transformed into their original format and passed on to the agent to continue the cycle.

Accordingly, for those forward models parameterised by bootstrapped multi-headed networks the last layer produces $K$ frames and rewards. As outlined in sections \ref{section:bootstrap} and \ref{section:btf}, RPFs require a prior in the form of an additional neural network with fixed parameters. The prior network maintains the same structure, inputs and outputs of the bootstrapped network.

\section{Results}

\begin{figure*}
     \includegraphics[width=\textwidth]{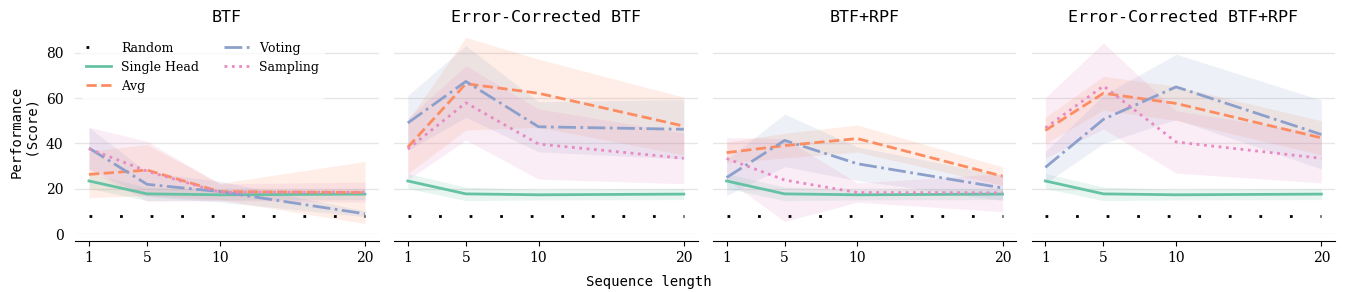}
     \centering
     \vspace*{-6mm}
     \caption{Comparison of uncertainty-aware BTF agents against a non-probabilistic agent with a single-headed model. The plots show the performance in Minipacman with different planning horizons.}
     \label{fig:performance}
\end{figure*}

\subsection{Model learning}

\begin{table}[h]
    \begin{adjustbox}{width=\columnwidth,center}
      \begin{tabular}{ c  c  c  c  c  c  c  c  }
        \multicolumn{8}{c}{\textbf{Accuracy}} \\
      \toprule
      \midrule      
      & \emph{\textbf{Single Head}} & \multicolumn{2}{ c }{\emph{\textbf{Average}}} & \multicolumn{2}{c}{\emph{\textbf{Majority Voting}}}& \multicolumn{2}{c}{\emph{\textbf{Sampling}}} \\
      & & \emph{\textbf{Boot}} & \emph{\textbf{RPF}} & \emph{\textbf{Boot}} & \emph{\textbf{RPF}} & \emph{\textbf{Boot}} & \emph{\textbf{RPF}}\\
      \midrule
       Fruit & 0.99583 & 0.99627 & 0.99617 & 0.99627 & \textbf{0.99638} & 0.99617 & 0.99617\\
       Eaten Cell & \textbf{0.99598} & 0.99545 & 0.99261 & 0.99564 & 0.9926 & 0.99506 & 0.99174\\
       Pacman & 0.90284 & 0.99899 & 0.99899 & 0.99919 & \textbf{0.99939} & 0.99878 & 0.99777\\
       Pacman (EC) & - & 0.99939 & 0.99979 & 0.99939 & \textbf{0.99979} & 0.99939 & 0.99959\\
       Ghost & 0.22980 & 0.36043 & \textbf{0.47513} & 0.35595 & 0.46303 & 0.34722 & 0.40053\\
       Ghost (EC) & - & 0.49955 & 0.50134 & 0.49955 & \textbf{0.50179} & 0.49865 & \textbf{0.50179}\\
       \hline
       Reward & 0.9557 & 0.97356 & 0.97376 & \textbf{0.97519} & 0.97437 & 0.97112 & 0.97397\\
       Frame & 0.9962 & 0.99661 & 0.9966 & \textbf{0.99663} & 0.99661 & 0.9966 & 0.99658\\ 
       \midrule
       \bottomrule
        \multicolumn{8}{c}{\textbf{Constraint Fulfilment}} \\
      \toprule      
      \midrule
       Fruit & 0.86594 & 0.96433 & 0.98296 & 0.95127 & \textbf{0.98566} & 0.9548 & 0.97173\\
       Pacman & 0.90499 & 0.95564 & \textbf{0.99768} & 0.99654 & 0.99584 & 0.97429 & 0.99558\\
       Pacman (EC) & - & 1.0 & 1.0 & 1.0 & 1.0 & 1.0 & 1.0\\
       Ghost & 0.31941 & 0.63119 & 0.85846 & 0.63407 & \textbf{0.86749} & 0.51388 & 0.7348\\
       Ghost (EC) & - & 0.92741 & 0.9197 & 0.93 & \textbf{0.95096} & 0.94887 & 0.92379\\
       \bottomrule
    \end{tabular}
    \caption{Model learning. Top: Average next step accuracy by category and type of forward model. Bottom: The proportion of frames that satisfied exactly the constraints as observed in the environment.} \label{tab:results}
    \end{adjustbox}
\end{table}

The first empirical evaluation considers the capacity to predict future state and reward. We compared a non-probabilistic single-headed agent (SH) against six variations of a 10-headed CNN architecture as described in the previous section. Three of the agents operated with a forward model trained through bootstrapped transition functions (BTF) while the other three also included a random prior (BTF+RPF). For both groups we tested the three mechanisms to consolidate predictions defined by eq. \ref{eqn:avg}, \ref{eqn:vote} and \ref{eqn:sample}. All CNNs were trained for 50,000 steps using a random policy and then tested on their next step predictions over 100 episodes. The accuracy was measured by verifying the number of cells (or reward class) that matched with the ground truth. If we look only at the average accuracy in the entire frame, all models appear to be deceivingly close. However, decomposing the accuracy by element category it is easier to appreciate the differences that arise when predicting the two most dynamic elements of the game, namely the ghost and pacman (upper section table \ref{tab:results}). Uncertainty-aware agents are consistently able to predict the motor consequences of their actions to a larger extent than when using a SH model. However all agents are less adept at predicting the semi-stochastic behaviour of the ghost. Activating the error-correcting routines has a positive effect for both categories. They allow the agents to predict correctly the ghost movement up to 50\% of the time. 

Error-correcting mechanisms however provided something considerably more critical. From the observations we had gathered on model prediction it was not rare to find frames where pacman was missing or cloned. This was also a common occurrence with the ghost. The bottom section of table \ref{tab:results} refers to the proportion of frames that complied exactly with the element constraints imposed by the environment. That is, the predicted frames that did not miss or included extra elements compared to the real observation. There we can observe that error-correcting mechanisms rightly included pacman and increased significantly the likelihood of removing or including the exact number of ghosts that appear in the next time step.

\subsection{Planning and game performance}

We measured the performance of the agent in the game to analyse the impact that the probabilistic models have on planning with simulated temporal sequences. The models were tested in combination with RHE for sequences of 1, 5, 10 and 20 steps into the future, a population size of 10, a mutation rate of 0.9 and evaluated for 10 episodes. We observed that in general all bootstrapped models, given a choice of sequence length, offered better performance than the SH model. At their peak, bootstrapped models without error-correction achieved a 2x improvement and above 3x when enhanced with error-correction. All models degraded in performance as horizon depth increases, however error-correcting models tended to benefit more from longer sequences and managed to retain proficiency even with 20-step sequences. For reference, RHE using a perfect simulator achieves a mean score of 134.54.

We also noticed that BTF+RPF exhibits more stable performance than BTF alone. Supporting these observations with the results in model learning, we could draw parallels with \cite{osband2018} and speculate that the prior is indeed acting as a regularisation mechanism. This could imply that during training the ensemble is more diverse in its predictions and therefore less susceptible to premature overconfident estimations.

\section{Discussion and Future Work}

The original objective behind this work was to investigate potential benefits in performance when incorporating uncertainty estimation into the forward model of an agent. Although in the literature there has been a large emphasis on producing more accurate models, our results support the conclusions reached in previous work regarding the deficiency of pursuing blindly this approach \cite{talvitie2014}. Producing highly accurate immediate predictions may still lead to catastrophic failure if global accuracy measures and loss functions fail to capture how the essential components of the environment are modelled. Thus there is a necessity to study further which additional criteria should assist in evaluating the viability of a forward model. We focused on structural consistency in the predictions as one of these criteria, by relating them to the specific environmental constraints they should satisfy. We explored this notion by introducing post-processing error-correcting mechanisms and showed that it is possible to take advantage of the predictions from an ensemble to guide the corrections. 

The results demonstrated that training a forward model capable of producing uncertainty estimates leads to improved performance and that enhancing them with error-correction not only improves it further but also increases stability in longer horizons. On a conceptual level this may suggest that authenticity or frame coherence could be complementary to (granular) accuracy for correctly planning in uncertain environments. Because we were interested in the implications of uncertainty estimation with inaccurate forward models some of the experimental choices were intended to study the robustness of our approach. For example, using a high mutation rate to produce a widely diverse set of sequences to test the model or planning with an unbiased RHE. Thus it is possible that the agents could perform better on the task with adequate hyperparameter tuning and we could establish comparisons against similar methods such as I2A \cite{racaniere2017}.

Although BTF by itself can be generically applicable to multiple domains, BTF with error-correction is provided as a starting point and it is still at this moment limited in scope. There are several directions for future work. For instance, we have assumed prior knowledge of the constraints. One could imagine learning these constraints directly from data by predicting the presence of elements and passing on this prediction to the error-correcting routines. Another future challenge is to scale the error-correcting capacity beyond grid-based games as well as the possibility of developing other mechanisms able to operate in latent space. Lastly, a future study could consider the inclusion of a density estimator or a novelty detector for an additional assessment of the structural consistency and realism of a prediction. 

\section*{Acknowledgement}

This research utilised Queen Mary's Apocrita HPC facility, supported by QMUL Research-IT. \url{http://doi.org/10.5281/zenodo.438045}


\bibliographystyle{ieeetr}
\bibliography{refs}

\begin{thebibliography}{10}

\bibitem{buesing2018}
L.~Buesing, T.~Weber, S.~Racaniere, S.~M.~A. Eslami, D.~Rezende, D.~P.
  Reichert, F.~Viola, F.~Besse, K.~Gregor, D.~Hassabis, and D.~Wierstra,
  ``Learning and {{Querying Fast Generative Models}} for {{Reinforcement
  Learning}},'' Feb. 2018.

\bibitem{hafner2019}
D.~Hafner, T.~Lillicrap, I.~Fischer, R.~Villegas, D.~Ha, H.~Lee, and
  J.~Davidson, ``Learning {{Latent Dynamics}} for {{Planning}} from
  {{Pixels}},'' in {\em International {{Conference}} on {{Machine Learning}}},
  pp.~2555--2565, May 2019.

\bibitem{ha2018}
D.~Ha and J.~Schmidhuber, ``Recurrent {{World Models Facilitate Policy
  Evolution}},'' in {\em Advances in {{Neural Information Processing Systems}}
  31} (S.~Bengio, H.~Wallach, H.~Larochelle, K.~Grauman, N.~{Cesa-Bianchi}, and
  R.~Garnett, eds.), pp.~2450--2462, {Curran Associates, Inc.}, 2018.

\bibitem{asadi2019}
K.~Asadi, D.~Misra, S.~Kim, and M.~L. Littman, ``Combating the
  {{Compounding}}-{{Error Problem}} with a {{Multi}}-step {{Model}},'' {\em
  arXiv:1905.13320 [cs, stat]}, May 2019.

\bibitem{schrittwieser2020}
J.~Schrittwieser, I.~Antonoglou, T.~Hubert, K.~Simonyan, L.~Sifre, S.~Schmitt,
  A.~Guez, E.~Lockhart, D.~Hassabis, T.~Graepel, T.~Lillicrap, and D.~Silver,
  ``Mastering {{Atari}}, {{Go}}, {{Chess}} and {{Shogi}} by {{Planning}} with a
  {{Learned Model}},'' {\em arXiv:1911.08265 [cs, stat]}, Feb. 2020.

\bibitem{kendall2017}
A.~Kendall and Y.~Gal, ``What uncertainties do we need in {{Bayesian}} deep
  learning for computer vision?,'' in {\em Proceedings of the 31st
  {{International Conference}} on {{Neural Information Processing Systems}}},
  {{NIPS}}'17, ({Long Beach, California, USA}), pp.~5580--5590, {Curran
  Associates Inc.}, Dec. 2017.

\bibitem{racaniere2017}
S.~Racani{\`e}re, T.~Weber, D.~Reichert, L.~Buesing, A.~Guez,
  D.~Jimenez~Rezende, A.~Puigdom{\`e}nech~Badia, O.~Vinyals, N.~Heess, Y.~Li,
  R.~Pascanu, P.~Battaglia, D.~Hassabis, D.~Silver, and D.~Wierstra,
  ``Imagination-{{Augmented Agents}} for {{Deep Reinforcement Learning}},'' in
  {\em Advances in {{Neural Information Processing Systems}} 30} (I.~Guyon,
  U.~V. Luxburg, S.~Bengio, H.~Wallach, R.~Fergus, S.~Vishwanathan, and
  R.~Garnett, eds.), pp.~5690--5701, {Curran Associates, Inc.}, 2017.

\bibitem{efron1979}
B.~Efron, ``Bootstrap {{Methods}}: {{Another Look}} at the {{Jackknife}},''
  {\em The Annals of Statistics}, vol.~7, pp.~1--26, Jan. 1979.

\bibitem{osband2016a}
I.~Osband, C.~Blundell, A.~Pritzel, and B.~V. Roy, ``Deep exploration via
  bootstrapped {{DQN}},'' in {\em Proceedings of the 30th {{International
  Conference}} on {{Neural Information Processing Systems}}}, {{NIPS}}'16,
  ({Barcelona, Spain}), pp.~4033--4041, {Curran Associates Inc.}, Dec. 2016.

\bibitem{mnih2013}
V.~Mnih, K.~Kavukcuoglu, D.~Silver, A.~Graves, I.~Antonoglou, D.~Wierstra, and
  M.~Riedmiller, ``Playing {{Atari}} with {{Deep Reinforcement Learning}},''
  {\em arXiv:1312.5602 [cs]}, Dec. 2013.

\bibitem{osband2018}
I.~Osband, J.~Aslanides, and A.~Cassirer, ``Randomized {{Prior Functions}} for
  {{Deep Reinforcement Learning}},'' in {\em Advances in {{Neural Information
  Processing Systems}} 31} (S.~Bengio, H.~Wallach, H.~Larochelle, K.~Grauman,
  N.~{Cesa-Bianchi}, and R.~Garnett, eds.), pp.~8617--8629, {Curran Associates,
  Inc.}, 2018.

\bibitem{huang2019}
W.~Huang, J.~Zhang, and K.~Huang, ``Bootstrap {{Estimated Uncertainty}} of the
  {{Environment Model}} for {{Model}}-{{Based Reinforcement Learning}},'' {\em
  Proceedings of the AAAI Conference on Artificial Intelligence}, vol.~33,
  pp.~3870--3877, July 2019.

\bibitem{farquhar2018}
G.~Farquhar, T.~Rockt{\"a}schel, M.~Igl, and S.~Whiteson, ``{{TreeQN}} and
  {{ATreeC}}: {{Differentiable Tree}}-{{Structured Models}} for {{Deep
  Reinforcement Learning}},'' in {\em International {{Conference}} on
  {{Learning Representations}}}, Feb. 2018.

\bibitem{hinton1993}
G.~E. Hinton and D.~{van Camp}, ``Keeping the neural networks simple by
  minimizing the description length of the weights,'' in {\em Proceedings of
  the Sixth Annual Conference on {{Computational}} Learning Theory}, {{COLT}}
  '93, ({Santa Cruz, California, USA}), pp.~5--13, {Association for Computing
  Machinery}, Aug. 1993.

\bibitem{graves2011}
A.~Graves, ``Practical {{Variational Inference}} for {{Neural Networks}},'' in
  {\em Advances in {{Neural Information Processing Systems}} 24}
  (J.~{Shawe-Taylor}, R.~S. Zemel, P.~L. Bartlett, F.~Pereira, and K.~Q.
  Weinberger, eds.), pp.~2348--2356, {Curran Associates, Inc.}, 2011.

\bibitem{blundell2015}
C.~Blundell, J.~Cornebise, K.~Kavukcuoglu, and D.~Wierstra, ``Weight
  {{Uncertainty}} in {{Neural Network}},'' in {\em International {{Conference}}
  on {{Machine Learning}}}, pp.~1613--1622, June 2015.

\bibitem{perez2013}
D.~Perez, S.~Samothrakis, S.~Lucas, and P.~Rohlfshagen, ``Rolling horizon
  evolution versus tree search for navigation in single-player real-time
  games,'' in {\em Proceedings of the 15th Annual Conference on {{Genetic}} and
  Evolutionary Computation}, {{GECCO}} '13, ({Amsterdam, The Netherlands}),
  pp.~351--358, {Association for Computing Machinery}, July 2013.

\bibitem{liu2016}
J.~Liu, D.~{P{\'e}rez-Li{\'e}bana}, and S.~M. Lucas, ``Rolling {{Horizon
  Coevolutionary}} planning for two-player video games,'' in {\em 2016 8th
  {{Computer Science}} and {{Electronic Engineering}} ({{CEEC}})},
  pp.~174--179, Sept. 2016.

\bibitem{santos2018}
B.~Santos, H.~Bernardino, and E.~Hauck, ``An {{Improved Rolling Horizon
  Evolution Algorithm}} with {{Shift Buffer}} for {{General Game Playing}},''
  in {\em 2018 17th {{Brazilian Symposium}} on {{Computer Games}} and {{Digital
  Entertainment}} ({{SBGames}})}, pp.~31--316, Oct. 2018.

\bibitem{gaina2017}
R.~D. Gaina, S.~M. Lucas, and D.~{Perez-Liebana}, ``Rolling horizon evolution
  enhancements in general video game playing,'' in {\em 2017 {{IEEE
  Conference}} on {{Computational Intelligence}} and {{Games}} ({{CIG}})},
  pp.~88--95, Aug. 2017.

\bibitem{gaina2017a}
R.~D. Gaina, S.~M. Lucas, and D.~{P{\'e}rez-Li{\'e}bana}, ``Population seeding
  techniques for {{Rolling Horizon Evolution}} in {{General Video Game
  Playing}},'' in {\em 2017 {{IEEE Congress}} on {{Evolutionary Computation}}
  ({{CEC}})}, pp.~1956--1963, June 2017.

\bibitem{tong2019}
X.~Tong, W.~Liu, and B.~Li, ``Enhancing {{Rolling Horizon Evolution}} with
  {{Policy}} and {{Value Networks}},'' in {\em 2019 {{IEEE Conference}} on
  {{Games}} ({{CoG}})}, pp.~1--8, Aug. 2019.

\bibitem{kulkarni2019}
T.~Kulkarni, A.~Gupta, C.~Ionescu, S.~Borgeaud, M.~Reynolds, A.~Zisserman, and
  V.~Mnih, ``Unsupervised {{Learning}} of {{Object Keypoints}} for
  {{Perception}} and {{Control}},'' {\em arXiv:1906.11883 [cs]}, Nov. 2019.

\bibitem{watters2019}
N.~Watters, L.~Matthey, M.~Bosnjak, C.~P. Burgess, and A.~Lerchner,
  ``{{COBRA}}: {{Data}}-{{Efficient Model}}-{{Based RL}} through {{Unsupervised
  Object Discovery}} and {{Curiosity}}-{{Driven Exploration}},'' {\em
  arXiv:1905.09275 [cs]}, Aug. 2019.

\bibitem{anand2020}
A.~Anand, E.~Racah, S.~Ozair, Y.~Bengio, M.-A. C{\^o}t{\'e}, and R.~D. Hjelm,
  ``Unsupervised {{State Representation Learning}} in {{Atari}},'' {\em
  arXiv:1906.08226 [cs, stat]}, Jan. 2020.

\bibitem{talvitie2014}
E.~Talvitie, ``Model regularization for stable sample rollouts,'' in {\em
  Proceedings of the {{Thirtieth Conference}} on {{Uncertainty}} in
  {{Artificial Intelligence}}}, {{UAI}}'14, ({Quebec City, Quebec, Canada}),
  pp.~780--789, {AUAI Press}, July 2014.

\end{thebibliography}

\vspace{12pt}

\end{document}